\documentclass{article}

\usepackage{color}
\usepackage{amsmath}
\usepackage{amsthm}
\usepackage{amsfonts}
\usepackage{amssymb}
\usepackage{algorithm}
\usepackage{algorithmic}
\usepackage[pdftex]{graphicx} \pdfcompresslevel=9
\usepackage{wrapfig}
\usepackage{paralist}

\title{Portfolio Allocation for Bayesian Optimization}
\author{
Eric Brochu, Matthew Hoffman, Nando de Freitas\\
Department of Computer Science\\
University of British Columbia\\
Vancouver, Canada\\
\texttt{\{ebrochu,hoffmanm,nando\}@cs.ubc.ca}
}

\newcommand{\mbs}[1]{\ensuremath{\boldsymbol{#1}}}

\newcommand{\thetav}{\mbs{\theta}}

\newcommand{\data}{{\mathcal D}}

\newtheorem{lemma}{Lemma}
\newtheorem{theorem}{Theorem}

\newcommand{\kv}{\mathbf{k}}

\newcommand{\x}{\mathbf{x}}
\newcommand{\y}{\mathbf{y}}

\newcommand{\K}{\mathbf{K}}

\newcommand{\func}{(\cdot)}

\newcommand{\xstar}{\mathbf{x}^{\star}}
\newcommand{\xbest}{\mathbf{x}^{+}}

\DeclareMathOperator*{\argmax}{argmax}

\newcommand{\sfrac}[2]{\leavevmode\kern.1em
           \raise.5ex\hbox{\footnotesize #1}\kern-.1em
                   /\kern-.15em\lower.25ex\hbox{\footnotesize #2}}

\def\capstyle#1{\small \emph{#1}}

\def\sigman{\sigma}

\begin{document}
\maketitle
\begin{abstract}
    Bayesian optimization with Gaussian processes has become an increasingly popular tool in the machine learning community.  It is efficient and can be used when very little is known about the objective function, making it popular in expensive black-box optimization scenarios.  It uses Bayesian methods to sample the objective efficiently using an \emph{acquisition function} which incorporates the model's estimate of the objective and the uncertainty at any given point.  However, there are several different parameterized acquisition functions in the literature, and it is often unclear which one to use. Instead of using a single acquisition function, we adopt a portfolio of acquisition functions governed by an online multi-armed bandit strategy.
We propose several portfolio strategies, the best of which we call GP-Hedge,
and show that this method outperforms the best individual acquisition function. We also provide a theoretical bound on the algorithm's performance.
\end{abstract}

\section{Introduction}

\emph{Bayesian optimization} is a powerful strategy for finding the extrema of objective functions that are expensive to evaluate. It is applicable in situations where one does not have a closed-form expression for the objective function, but where one can obtain noisy evaluations of this function at sampled values. It is particularly useful when these evaluations are costly, when one does not have access to derivatives, or when the problem at hand is non-convex. Bayesian optimization has two key ingredients. First, it uses the entire sample history to compute a posterior distribution over the unknown objective function. Second, it uses an \emph{acquisition function} to automatically trade off between exploration and exploitation when selecting the points at which to sample next.     
As such, Bayesian optimization techniques are some of the most efficient approaches in terms of the number of function evaluations required \cite{Mockus:1978,Jones:1998,Lizotte:2008,Boyle:2007,Brochu:2010b}.  In recent years, the machine learning community has increasingly used Bayesian optimization to optimize expensive objective functions.  Examples can be found in robot gait design \cite{Lizotte:2007}, online path planning \cite{Martinez-Cantin:2009,Martinez-Cantin:2007}, 
intelligent user interfaces for animation \cite{Brochu:2007b,Brochu:2010}, algorithm configuration \cite{Hutter:2009a}, efficient MCMC \cite{Rasmussen:2003}, sensor placement \cite{Srinivas:2010,Osborne:2010}, and reinforcement learning \cite{Brochu:2010b}.

However, the choice of acquisition function is not trivial.  Several different methods have been proposed in the literature, none of which work well for all classes of functions. Building on recent developments in the field of online learning and multi-armed bandits \cite{Cesa-Bianchi:2006}, this paper proposes a solution to this problem. The solution is based on a hierarchical hedging approach for managing an adaptive portfolio of acquisition functions.

We review Bayesian optimization and popular acquisition functions in Section 2. In Section 3, we propose the use of various hedging strategies for Bayesian optimization \cite{Auer:1998,Chaudhuri:2009}. In Section 4, we present experimental results using standard test functions from the literature of global optimization. The experiments show that the proposed hedging approaches outperform any of the individual acquisition functions. We also provide detailed comparisons among the hedging strategies. 
Finally, in Section 5 we present a bound on the cumulative regret which helps provide some intuition as to algorithm's performance.


\section{Bayesian optimization}\label{sec:background}
We are concerned with the task of optimization on a $d$-dimensional space:
$
\max_{\x \in A \subseteq \mathbb{R}^d} f(\x).
$

We define $\x_t$ as the $t$th sample and $y_t = f(\x_t) + \epsilon_t,$ with $\epsilon_t \stackrel{iid}{\sim} \mathcal{N}(0, \sigman^2)$, as a noisy observation of the objective function at $\x_t$. Other observation models are possible \cite{Brochu:2010b,Chu:2005a,Diggle1998,Rue2009}, but we will focus on real, Gaussian observations for ease of presentation. 

The Bayesian optimization procedure is shown in Algorithm~\ref{alg1}. As mentioned earlier, it has two components: the posterior distribution over the objective and the acquisition function. Let us focus on the posterior distribution first and come back to the acquisition function in Section~\ref{sec:acquisition}.
As we accumulate observations\footnote{
Here we use subscripts to denote sequences of data, i.e. $y_{1:t}=\{y_1,\dots,y_t\}$.
} $\data_{1:t} = \{\x_{1:t},y_{1:t}\}$, a prior distribution $P(f)$ is combined with the likelihood function $P(\data_{1:t}|f)$ 
to produce the posterior distribution:
$
P(f|\data_{1:t}) \propto P(\data_{1:t}|f) P(f).
$
The posterior captures the updated beliefs about the unknown objective function.
One may also interpret this step of Bayesian optimization as estimating the objective function with a \emph{surrogate function} (also called a \emph{response surface}). We will place a Gaussian process (GP) prior on $f$. Other nonparametric priors over functions, such as random forests, have been considered \cite{Brochu:2010b}, but the GP strategy is the most popular alternative.

\begin{algorithm}
\caption{Bayesian Optimization}\label{alg:bayopt}
\begin{algorithmic}[1]
{\footnotesize
   \FOR{$t=1,2,\dots$}
       \STATE Find $\x_t$ by optimizing the acquisition function over the GP: $\x_t = \argmax_{\x}u(\x|\data_{1:t-1})$.
       \STATE Sample the objective function: $y_t=f(\x_t)+\epsilon_t$.
       \STATE Augment the data $\data_{1:t} = \{\data_{1:t-1}, (\x_t, y_t)\}$. 	
   \ENDFOR
}
\label{alg1}
\end{algorithmic}
\end{algorithm}

\subsection{Gaussian processes}\label{sec:gp}
The objective function is distributed according to a GP prior:
\[
f(\x) \sim \operatorname{GP}(m(\x), k(\x_i,\x_j)).
\]
For convenience, and without loss of generality, we assume that the prior mean is the zero function (but see \cite{Martinez-Cantin:2007,Rasmussen:2006,Brochu:2010} for examples of nonzero means).  This leaves us the more interesting question of defining the covariance function.  A very popular choice is the squared exponential kernel with a vector of automatic relevance determination (ARD) hyperparameters $\thetav$ \cite{Rasmussen:2006}:
\[
k(\x_i, \x_j) = \exp\big(-\tfrac{1}{2}(\x_i-\x_j)^{T}
\operatorname{diag}(\thetav)^{-2}(\x_i-\x_j)\big), 
\]
where $\operatorname{diag}(\thetav)$ is a diagonal matrix with entries $\thetav$ along the diagonal and zeros elsewhere. The choice of hyperparameters will be discussed in the experimental section, but we note that it is not trivial in this domain because of the paucity of data. For an in depth analysis of this issue we refer the reader to e.g. \cite{Brochu:2010,Osborne:2010}.

We can sample the GP at $t$ points by choosing the indices $\{\x_{1:t}\}$ and sampling the values of the function at these indices to produce the data $\data_{1:t}$. The function values are distributed according to a multivariate Gaussian distribution $\mathcal{N}(0,\K)$, with covariance entries $ k(\x_i, \x_j)$.
Assume that we already have the observations, say from previous iterations, and that we want to use Bayesian optimization to decide what point $\x_{t+1}$ should be considered next. Let us denote the value of the function at this arbitrary point as $f_{t+1}$. Then, by the properties of GPs, $f_{1:t}$ and $f_{t+1}$ are jointly Gaussian:
\[
\begin{bmatrix}
    f_{1:t} \\
    f_{t+1}
\end{bmatrix}
\sim {\cal N} 
\left( \mathbf{0} ,
    \begin{bmatrix}
        \K & \kv \\
        \kv^{T} & k(\x_{t+1},\x_{t+1})
    \end{bmatrix}
\right),
\]
where 
$\kv = [k(\x_{t+1},\x_1), k(\x_{t+1},\x_2), \dots, k(\x_{t+1},\x_t)]$. 
Using the Sherman-Morrison-Woodbury formula, see \cite{Rasmussen:2006} for a comprehensive treatment, one can easily arrive 
at an expression for the predictive distribution:
\[
P(y_{t+1}|\data_{1:t},\x_{t+1}) = {\cal N} (\mu_t(\x_{t+1}), \sigma_t^2(\x_{t+1})+ \sigman^2),
\]
where
\begin{align*}
\mu_t(\x_{t+1})&=
\mathbf{k}^T [\mathbf{K} + \sigman^2\mathbf{I} ]^{-1} \y_{1:t}, \\
\sigma_t^2(\x_{t+1})&= k(\x_{t+1},\x_{t+1}) - \mathbf{k}^T [\mathbf{K} + \sigman^2\mathbf{I} ]^{-1}\mathbf{k}.
\nonumber
\end{align*}
In this sequential decision making setting, the number of query points is relatively small and, consequently, the GP predictions are easy to compute.

\begin{figure}[t!]
 \begin{center}
   \includegraphics[width=.8\columnwidth]{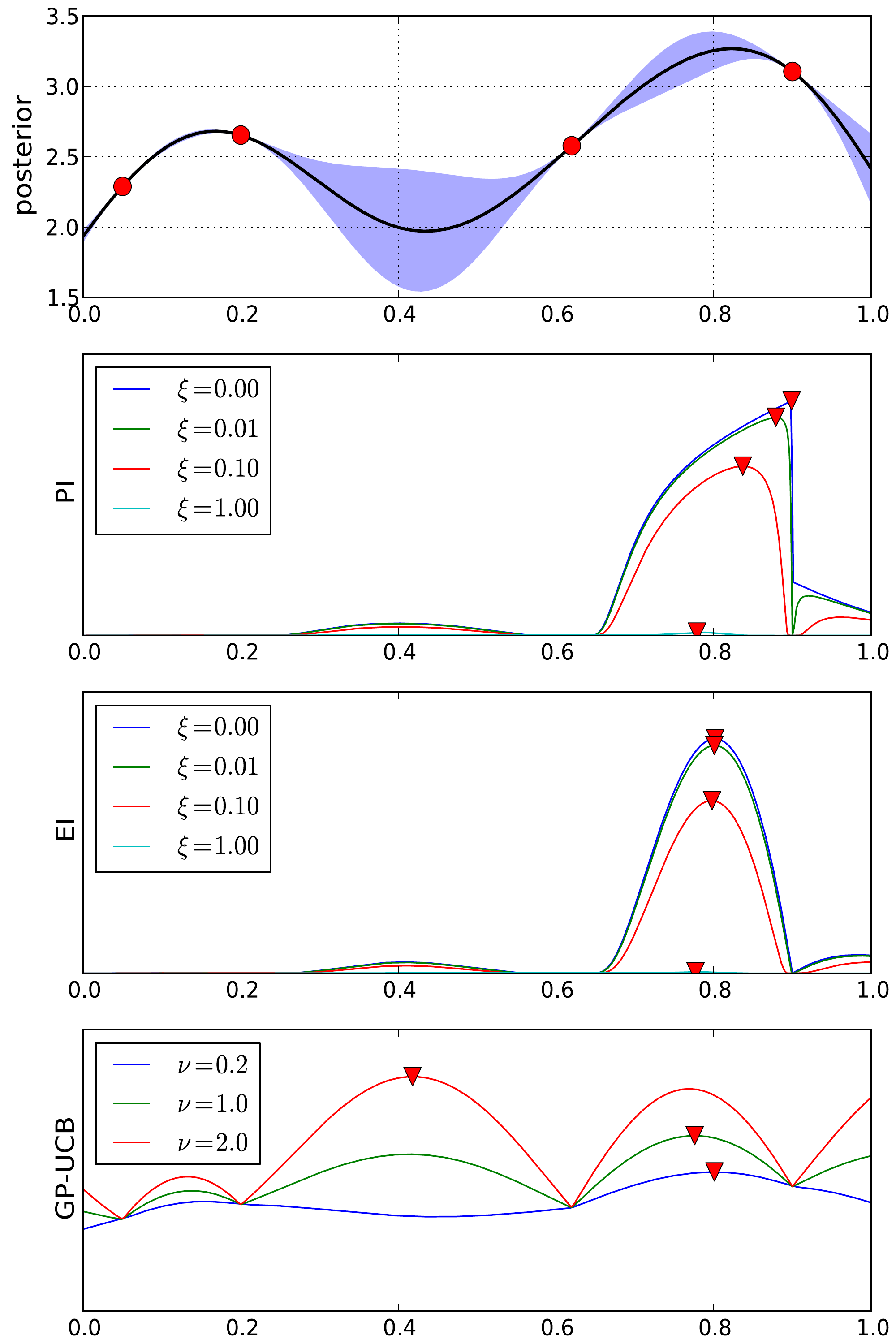}
 \end{center}
 \caption{\capstyle{Acquisition functions with different values of the exploration parameters $\nu$ and $\xi$.  The GP posterior is shown at the top.  The other images show the acquisition functions for that GP.  From the top: Probability of improvement, expected improvement and upper confidence bound.  The maximum of each acquisition function, where the GP is to be sampled next, is shown with a triangle marker. Note the increased preference for exploration exhibited by GP-UCB.}} 
 \label{fig:acquisition}
\end{figure}

\subsection{Acquisition functions}\label{sec:acquisition}
The role of the acquisition function is to guide the search for the optimum.  Typically, acquisition functions are defined such that high values correspond to \emph{potentially} high values of the objective function, whether because the prediction is high, the uncertainty is great, or both.  The acquisition function is maximized to select the next point at which to evaluate the objective function.  That is, we wish to sample the objective function at $\argmax_\x u(\x|\data)$. This auxiliary maximization problem, where the objective is known and easy to evaluate, can be easily carried out with standard numerical techniques such as multistart or DIRECT \cite{Jones:1993,Gablonsky:2001}.
The acquisition function is sometimes called the \emph{infill} or simply the ``utility'' function.  In the following sections, we will look at the three most popular choices.  Figure~\ref{fig:acquisition} shows how these give rise to distinct sampling behaviour.

\textbf{Probability of improvement (PI):}
The early work of Kushner \cite{Kushner:1964} suggested maximizing the \emph{probability of improvement} over the incumbent $\mu^+ = \max_t\mu(\x_t)$.
The drawback, intuitively, is that this formulation is biased toward exploitation only. To remedy this, practitioners often add a trade-off parameter $\xi \geq 0$, so that
\begin{align*}
\operatorname{PI}(\x) 
&= P(f(\x) \geq \mu^+ +\xi) 
= \Phi\left(\frac{\mu(\x) - \mu^+ - \xi}{\sigma(\x)}\right),
\end{align*}
where $\Phi\func$ is the standard Normal cumulative distribution function (CDF).  The exact choice of $\xi$ is left to the user. Kushner recommends using a (unspecified) schedule for $\xi$, which should start high in order to drive exploration and decrease towards zero as the algorithm progresses.  Lizotte, however, found that using such a schedule did not offer improvement over a constant value of $\xi$ on a suite of test functions \cite{Lizotte:2008}.

\textbf{Expected improvement (EI):}
More recent work has tended to take into account not only the probability of improvement, but the magnitude of the improvement a point can potentially yield.  Mo\v{c}kus et al.\ \cite{Mockus:1978} proposed maximizing the \emph{expected improvement} with respect to the best function value yet seen, given by the incumbent $\xbest = \argmax_{\x_t}f(\x_t)$. 
For our Gaussian process posterior, one can easily evaluate this expectation, see
\cite{Jones:2001}, yielding:
\begin{align*}
	\operatorname{EI}(\x) &= 
	\begin{cases}
		d\Phi(d/\sigma(\x)) + \sigma(\x) 
		\phi(d/\sigma(\x)) & \textrm{if } \sigma(\x) > 0\\
		0 & \textrm{if } \sigma(\x) = 0
	\end{cases}
\end{align*}
where $d = \mu(\x) - \mu^+ - \xi$ and where $\phi\func$ and $\Phi\func$ denote the PDF and CDF of the standard Normal distribution respectively. Here $\xi$ is an optional trade-off parameter analogous to the one defined above.

\textbf{Upper confidence bound (UCB \& GP-UCB):} Cox and John \cite{Cox:1997} introduce an algorithm they call ``Sequential Design for Optimization'', or \emph{SDO}.  Given a random function model, SDO selects points for evaluation based on a confidence bound consisting of the mean and weighted variance: $\mu(\x) + \kappa \sigma(\x)$.
As with the other acquisition models, however, the parameter $\kappa$ is left to the user. A principled approach to selecting this parameter is proposed by Srinivas et al.\ \cite{Srinivas:2010}. In this work, the authors define the instantaneous regret of the selection algorithm as $r(\x) = f(\xstar)-f(\x)$ and attempt to select a sequence of weights $\kappa_t$ so as to minimize the cumulative regret $R_T=r(\x_1)+\cdots+r(\x_T)$. 
Using the \emph{upper confidence bound} selection criterion with $\kappa_t=\sqrt{\nu\beta_t}$ and the hyperparameter $\nu>0$ Srinivas et al.\ define
\[
\operatorname{GP-UCB}(\x) = \mu(\x) + \sqrt{\nu\beta_t}\sigma(\x).
\]
It can be shown that this method has cumulative regret bounded by $\mathcal O(\sqrt{T\beta_T\gamma_T})$ with high probability. Here $\beta_T$ is a carefully selected learning rate and $\gamma_T$ is a bound on the information gained by the algorithm at selected points after $T$ steps. Both of these terms depend upon the particular form of kernel-function used, but for most kernels their product can be shown to be sublinear in $T$. We refer the interested reader to the original paper \cite{Srinivas:2010} for exact bounds.

The sublinear bound on cumulative regret implies that the method is \emph{no-regret}, i.e.\ that $\lim_{T\to\infty}R_T/T=0$. This in turn provides a bound on the convergence rate for the optimization process, since the regret at the maximum $f(\x^*) - \max_t f(\x_t)$ is upper bounded by the average regret 
$
R_T/T = 
f(\x^*) - \tfrac1T \textstyle{\sum_{t=1}^T} f(\x_t).
$ As we will note later, however, this bound can be quite loose in practice.


\begin{algorithm}
\caption{GP-Hedge}
\begin{algorithmic}[1]
{\footnotesize
    \STATE Select parameter $\eta \in \mathbb{R}^+$.
    \STATE Set $g^i_0 = 0$ for $i = 1, \ldots, N$.
    \FOR{$t=1,2,\dots$}
        \STATE Nominate points from each acquisition function: $\x_t^i = \argmax_\x u_i(\x|\data_{1:t-1})$.
        \STATE Select nominee $\x_t=\x_t^j$ with probability
        $p_t(j) = \exp(\eta g^j_{t-1}) / \sum_{\ell=1}^k \exp(\eta g^\ell_{t-1})$.
        \STATE Sample the objective function $y_t=f(\x_t)+\epsilon_t$.
        \STATE Augment the data $\data_{1:t} = \{\data_{1:t-1}, (\x_t, y_t)\}$. 
        \STATE Receive rewards $r_t^i = \mu_t(\x_t^i)$ from the updated GP.
        \STATE Update gains $g_t^i = g_{t-1}^i + r_t^i$. 
    \ENDFOR
}
\end{algorithmic}\label{alg:hedge}
\end{algorithm}

\section{Portfolio strategies}\label{sec:hedging}

There is no choice of acquisition function that can be guaranteed to perform best on an arbitrary, unknown objective.
In fact, it may be the case that no single acquisition function will perform the best over an entire optimization --- a mixed strategy in which the acquisition function samples from a pool (or portfolio) at each iteration might work better than any single acquisition. This can be treated as a hierarchical multi-armed bandit problem, in which each of the $N$ arms is an acquisition function, each of which is itself an infinite-armed bandit problem. In this section we propose solving the selection problem using three strategies from the literature, the application of which we believe to be novel.

\emph{Hedge} is an algorithm which at each time step $t$ selects an action $i$ with probability $p_t(i)$ based on the cumulative rewards (gain) for that action (see Auer et al.~\cite{Auer:1998}). After selecting an action the algorithm receives reward $r_t^i$ for each action and updates the gain vector. In the Bayesian optimization setting, we can define $N$ bandits each corresponding to a single acquisition function. Choosing action $i$ corresponds to sampling from the point nominated by function $u_i$, i.e.\ $\x_t^i = \argmax_\x u_i(\x|\data_{1:t-1})$ for $i = 1, \ldots, N$. Finally, while in the conventional Bayesian optimization setting the objective function is sampled only once per iteration, Hedge is a full information strategy and requires a reward for every action at every time step.  We can achieve this by defining the reward at $\x_t^i$ as the expected value of the GP model at $\x_t^i$.  That is, $r_t^i = \mu_t(\x^i_t)$. We refer to this method as GP-Hedge. Provided that the objective function is smooth, this reward definition is reasonable.

Auer et al.\ also propose the \emph{Exp3} algorithm, a variant of Hedge that applies to the partial information setting. In this setting it is no longer assumed that rewards are observed for all actions. Instead at each iteration a reward is only associated with the particular action selected. The algorithm uses Hedge as a subroutine where rewards observed by Hedge at each iteration are $r_t^i/\hat p_t(i)$ for the action selected and zero for all actions. Here $\hat p_t(i)$ is the probability that Hedge would have selected action $i$. The Exp3 algorithm, meanwhile, selects actions from a distribution that is a mixture between $\hat p_t(i)$ and the uniform distribution. Intuitively this ensures that the algorithm does not miss good actions because the initial rewards were low (i.e.\ it continues exploring).

Finally, another possible strategy is the \emph{NormalHedge} algorithm \cite{Chaudhuri:2009}. This method, however, is built to take advantage of situations where the number of bandit arms (acquisition functions) is large, and may not be a good match to problems where $N$ is relatively small.

The GP-Hedge procedure is shown in Algorithm~\ref{alg:hedge}.  In practice any of these hedging strategies could be used, however in our experiments we find that Hedge tends to outperform the others. Note that it is necessary to optimize $N$ acquisition functions at each time step rather than 1.  While this might seem expensive, this is unlikely to be a major problem in practice for small $N$, as (i) Bayesian optimization is typically employed when sampling the objective is so expensive as to dominate other costs; (ii) it has been shown that fast approximate optimization of $u$ is usually sufficient \cite{Brochu:2007b,Lizotte:2008,Hutter:2009a}; and (iii) it is straightforward to run the optimizations in parallel on a modern multicore processor.

We will also note that the setting of our problem is somewhere ``in between'' the full and partial information settings. Consider, for example, the situation that all points sampled by our algorithm are ``too distant'' in the sense that the kernels evaluated at these points exert negligible influence on each other. In this case, we can see that only information obtained by the sampled point is available, and as a result GP-Hedge will be over-confident in its predictions when using the full-information strategy. However, this behaviour is not observed in practical situations because of smoothness properties, as well as our particular selection of acquisition functions. In the case of adversarial acquisition functions one might instead choose to use the Exp3 variant.

\begin{figure*}[t!]
	\centering
	\includegraphics[width=\textwidth]{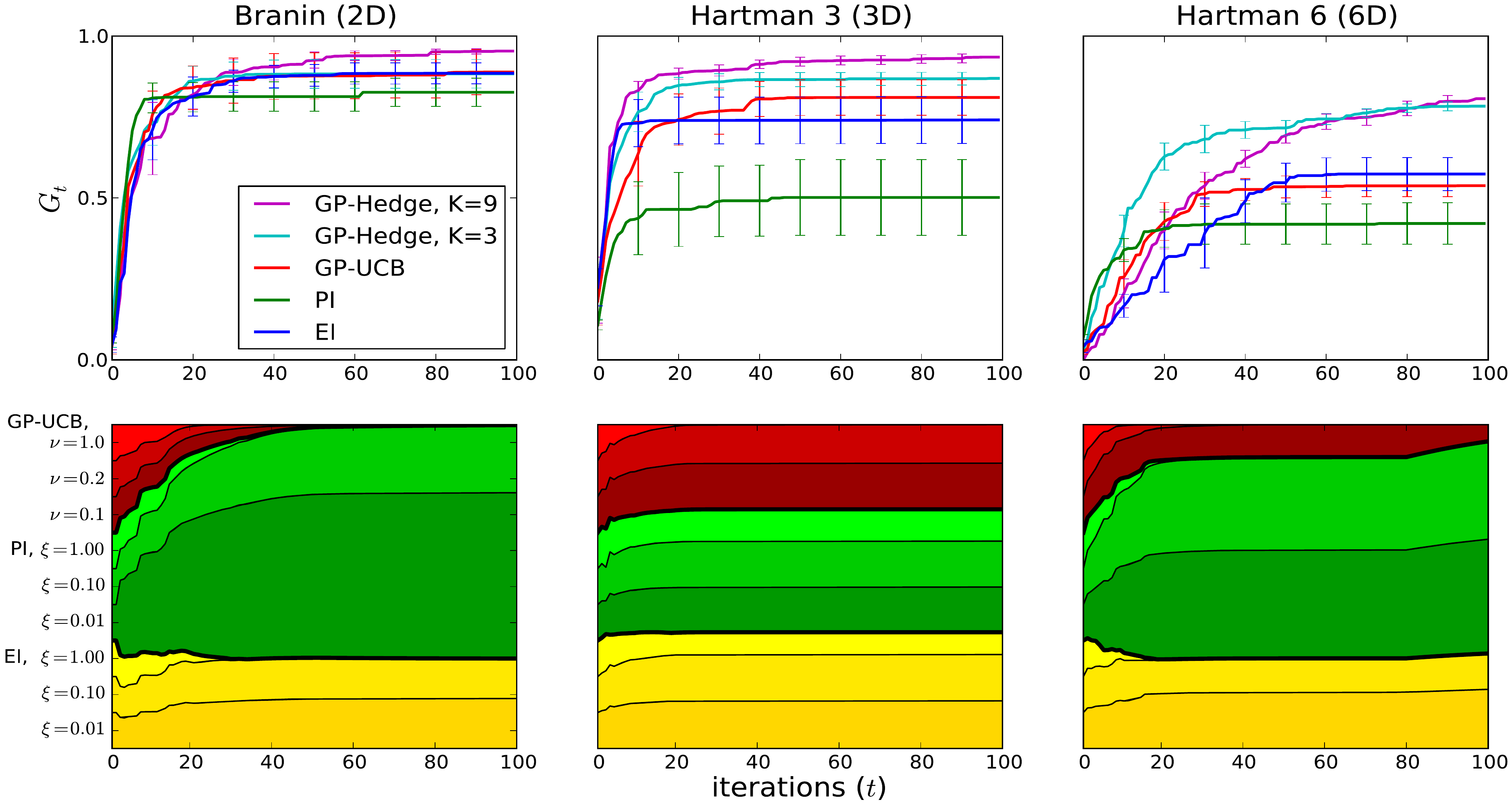}
	\caption{\capstyle{(Best viewed in colour.) Comparison of different acquisition approaches on three commonly used literature functions. The top plots show the mean and variance of the gap metric averaged over 25 trials.
We note that the top two performing algorithms use a portfolio strategy.
With $N=3$ acquisition functions, GP-Hedge beats the best-performing acquisition function in almost all cases.  With $N=9$, we add additional instances of the three acquisition functions, but with different parameters.  Despite the fact that these additional functions individually perform worse than the ones with default parameters, adding them to GP-Hedge improves performance in the long run. The bottom plots show an example evolution of GP-Hedge's portfolio with $N=9$ for each objective function. The height of each band corresponds to the probability $p_t(i)$ at each iteration.}}
 \label{fig:litfunc-gap}
\end{figure*}

\begin{figure*}[t!]
	\centering
	\includegraphics[width=\textwidth]{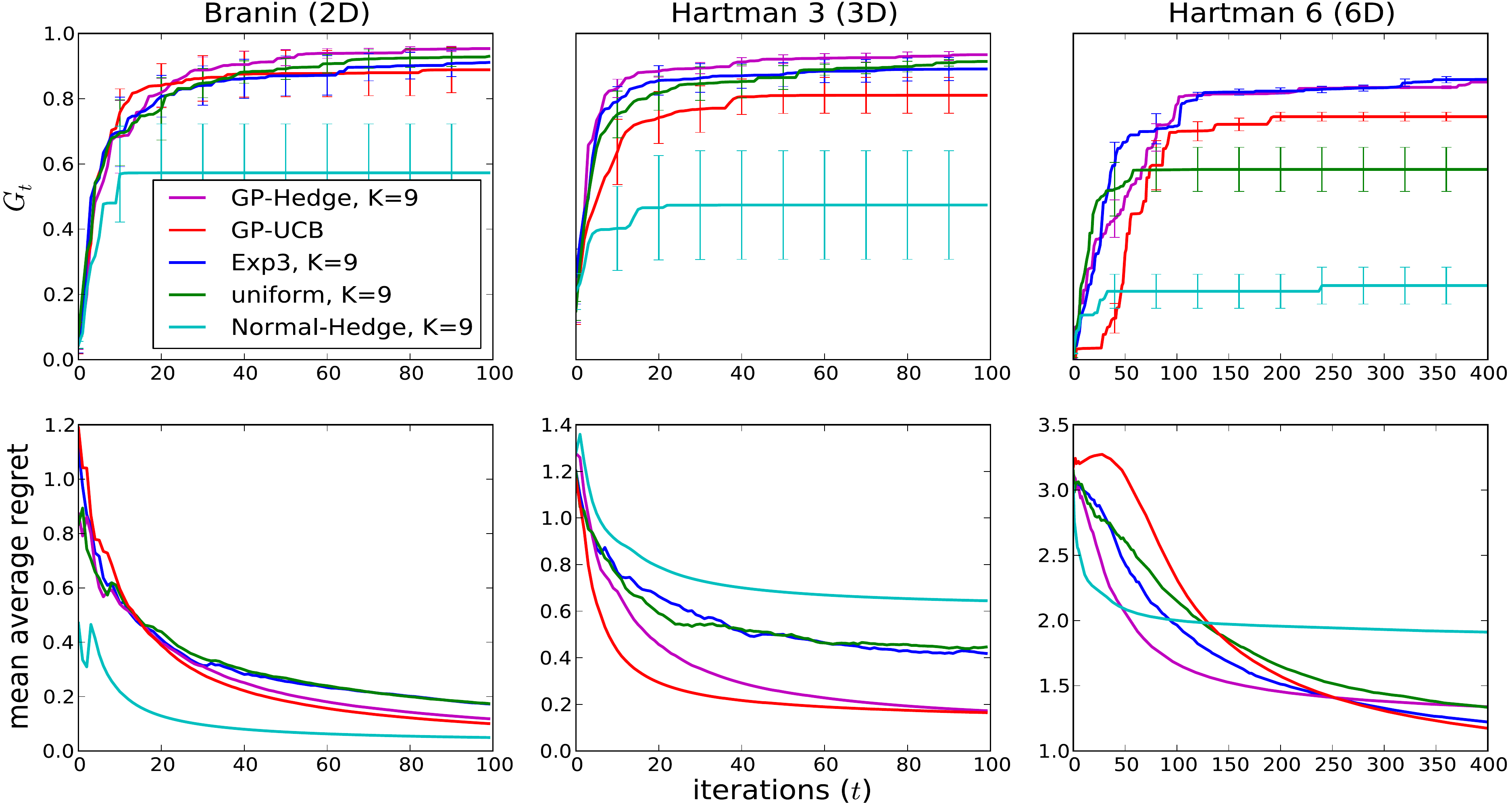}
	\caption{\capstyle{(Best viewed in colour.) Comparison of different hedging strategies on three commonly used literature functions. The top plots show the mean and variance of the gap metric averaged over 25 trials. Note that both Hedge and Exp3 outperform the best single acquisition function, GP-UCB.
The bottom plots show the mean average regret for each method (lower is better). Average regret is shown in order to compare with previous work~\cite{Srinivas:2010}, however as noted in the text the gap measure provides a more direct comparison of optimization performance. We see that mixed strategies (i.e.\ GP-Hedge) perform comparably to GP-UCB under the regret measure and outperform this individual strategy under the gap measure. As the problems get harder, and with higher dimensionality, GP-Hedge significantly outperforms other acquisition strategies.}}
 \label{fig:litfunc-regret}
\end{figure*}

\section{Experiments}

To validate the use of GP-Hedge, we tested the optimization performance on a set of test functions with known maxima $f(\x^\star)$\footnote{Code for the optimization methods and experiments will be made available online.}.  To see how effective each method is at finding the global maximum, we use the ``gap'' metric \cite{Huang:2006}, defined as
\[
G_t = \Big[f(\x^+)-f(\x_1)\Big] \Big/ \Big[f(\x^\star)-f(\x_1)\Big],
\]
where again $\xbest$ is the incumbent or best function sample found up to time $t$.
The gap $G_t$ will therefore be a number between 0 (indicating no improvement over the initial sample)
and 1 (if the incumbent is the maximum).
Note, while this performance metric is evaluated on the true function values, this information is not available to the optimization methods.

\subsection{Standard test functions}

We first tested performance using functions common to the literature on Bayesian optimization: the Branin, Hartman 3, and Hartman 6 functions.  All of these are continuous, bounded, and multimodal, with 2, 3, and 6 dimensions respectively.  We omit the formulae of the functions for space reasons, but they can be found in \cite{Lizotte:2008}.

For each experiment, we optimized 25 times and computed the mean and variance of the gap metric over time. In these experiments we used hyperparameters $\thetav$ chosen offline so as to maximize the log marginal likelihood of a (sufficiently large) set of sample points; see~\cite{Rasmussen:2006}.
We compared the standard acquisition functions using parameters suggested by previous authors, i.e.\ $\xi=0.01$ for EI and PI, $\delta=0.1$ and $\nu = 0.2$ for GP-UCB \cite{Lizotte:2008,Srinivas:2010}.
For the GP-Hedge trials, we tested performance under using both 3 acquisition functions and 9 acquisition functions.  For the 3-function variant we use the standard acquisition functions with default hyperparameters. The 9-function variant uses these same three as well as 6 additional acquisition functions consisting of: both PI and EI with $\xi=0.1$ and $\xi=1.0$, GP-UCB with $\nu = 0.1$ and $\nu = 1.0$.  While we omit trials of these additional acquisition functions for space reasons, these values are not expected to perform as well as the defaults and our experiments confirmed this hypothesis.  However, we are curious to see if adding known suboptimal acquisition functions will help or hinder GP-Hedge in practice.

Results for the gap measure $G_t$ are shown in Figure~\ref{fig:litfunc-gap}.  While the improvement GP-Hedge offers over the best single acquisition function varies, there is almost no combination of function and time step in which the 9-function GP-Hedge variant is not the best-performing method. The results suggest that the extra acquisition functions assist GP-Hedge in exploring the space in the early stages of the optimization process. Figure~\ref{fig:litfunc-gap} also displays, for a single example run, how the the arm probabilities $p_t(i)$ used by GP-Hedge evolve over time. We have observed that the distribution becomes more stable when the acquisition functions come to a general consensus about the best region to sample.  As the optimization progresses, exploitation becomes more rewarding than exploration, resulting in more probability being assigned to methods that tend to exploit. However, note that if the initial portfolio had consisted only of these more exploitative acquisition functions, the likelihood of becoming trapped at suboptimal points would have been higher.

In Figure~\ref{fig:litfunc-regret} we compare against the other Hedging strategies introduced in Section~\ref{sec:hedging} under both the gap measure and mean average regret. We also introduce a baseline strategy which utilizes a portfolio uniformly distributed over the same acquisition functions. The results show that mixing across multiple acquisition functions provides significant performance benefits under the gap measure, and as the problems' difficulty/dimensionality increases we see that GP-Hedge outperforms other mixed strategies. The uniform strategy performs well on the easier test functions, as the individual acquisition functions are reasonable.  However, for the hardest problem (Hartman 6) we see that the performance of the naive uniform strategy degrades. NormalHedge performs particularly poorly on this problem. We observed that this algorithm very quickly collapses to an exclusively exploitative portfolio which becomes very conservative in its departures from the incumbent. We again note that this strategy is intended for large values of $N$, which may explain this behaviour.

In the case of the regret measure we see that the hedging strategies perform comparable to GP-UCB, a method designed to optimize this measure. We further note that although the average regret can be seen as a lower-bound on the convergence of Bayesian optimization methods, this bound can be loose in practice. Further, in the setting of Bayesian optimization we are typically concerned not with the cumulative regret during optimization, but instead with the regret incurred by the incumbent after optimization is complete. Similar notions of ``simple regret'' have been studied in \cite{Audibert:2010,Bubeck:2009}.

Based on the performance in these experiments, we use Hedge as the underlying algorithm for GP-Hedge in the remainder of the experiments.



\begin{figure*}
 \begin{center}
   \includegraphics[width=\textwidth]{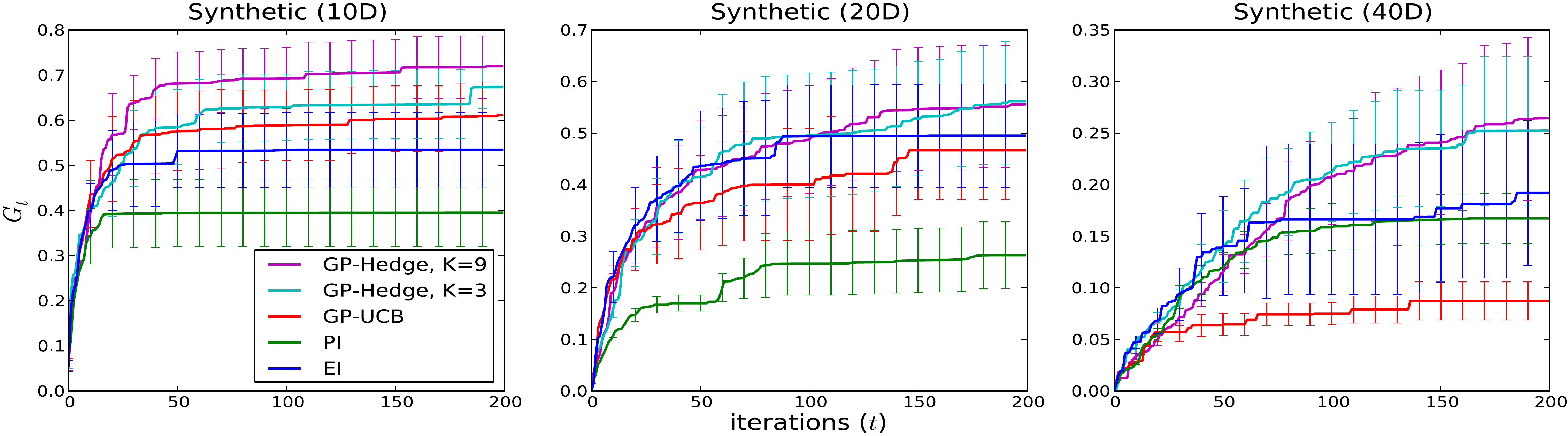}
 \end{center}
 \caption{\capstyle{(Best viewed in colour.) We compare the performance of the acquisition approaches on synthetic functions sampled from a GP prior with randomly initialized hyperparameters. Shown are the mean and variance of the gap metric over 25 sampled functions. Here, the variance is a relative measure of how well the various algorithms perform while the functions themselves are varied. While the variance is high (which is to be expected over diverse functions), we can see that GP-Hedge is at least comparable to the best acquisition functions and ultimately superior for both $N=3$ and $N=9$. We also note that for the 10D and 20D experiments GP-UCB performs quite well but suffers in the 40D experiment. This helps to confirm our hypothesis that a mixed strategy is particularly useful in situations where we do not possess strong prior information with regards to the choice of acquisition function.}}
 \label{fig:synfunc}
\end{figure*}

\subsection{Sampled test functions}

As there is no generally-agreed-upon set of test functions for Bayesian optimization in higher dimensions, we seek to sample synthetic functions from a known GP prior similar to \cite{Lizotte:2008}. For further details on how these functions are sampled see Appendix~\ref{sec:synthetics}.  As can be seen in Figure~\ref{fig:synfunc}, GP-Hedge with $N=9$ is again the best-performing method, which becomes even more clear as the dimensionality increases.  Interestingly, the \emph{worst}-performing function changes as dimensionality increases.  In the 40D experiments, GP-UCB, which generally performed well in other experiments, does quite poorly.  Examining the behaviour, it appears that by trying to minimize regret instead of maximizing improvement, GP-UCB favours regions of high variance.  However, since a 40D space remains extremely sparsely populated even with hundreds of samples, the vast majority of the space still has high variance, and thus high acquisition value.

\begin{figure}[b!]
 \begin{center}
   \includegraphics[width=\columnwidth]{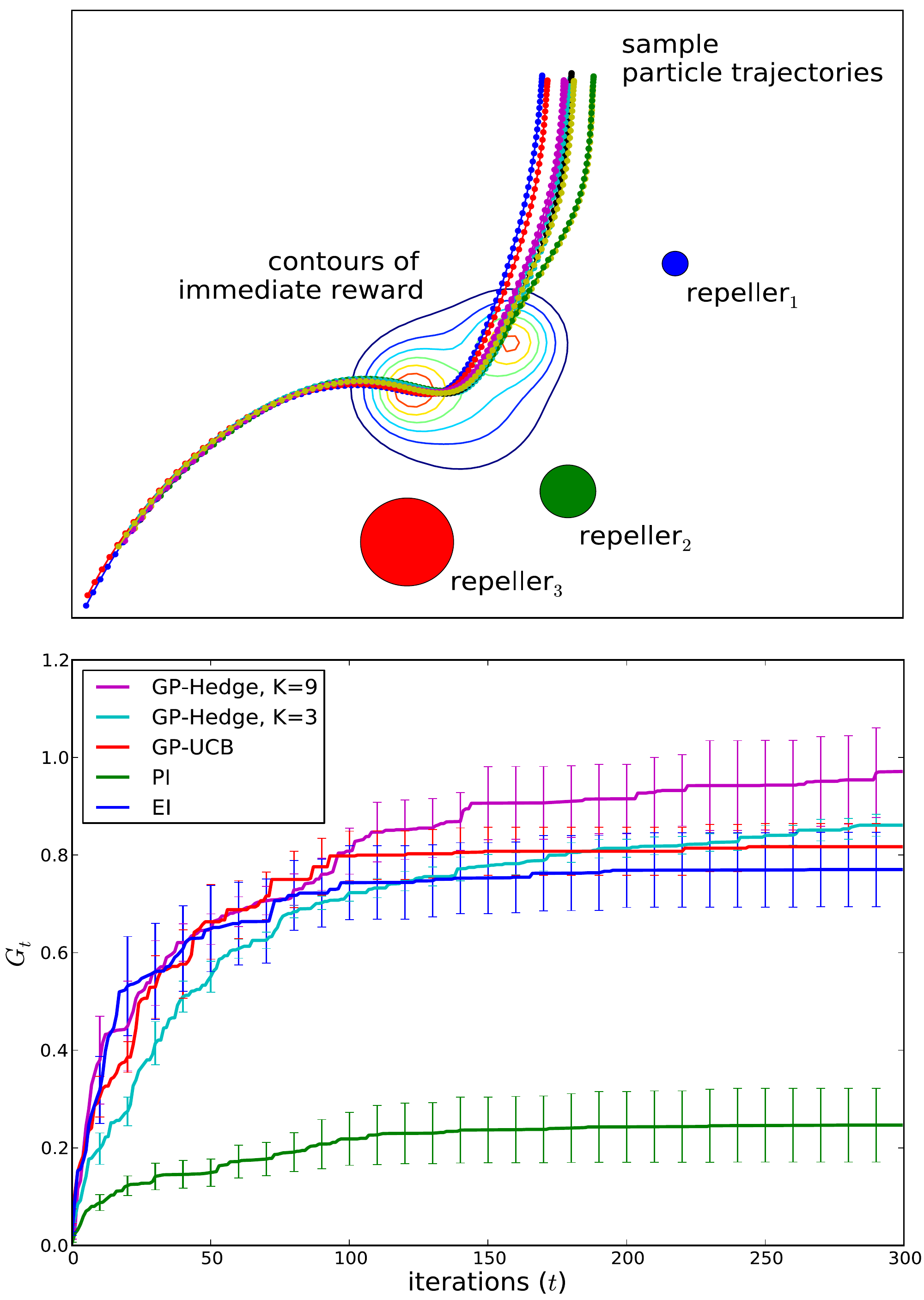}
 \end{center}
 \caption{\capstyle{(Best viewed in colour.) Results of experiments on the repeller control problem. The left-most plot displays 10 sample trajectories over 100 time-steps for a particular repeller configuration (not necessarily optimal). The right-most plot shows the progress of each of the described Bayesian optimization methods on a similar model, averaged over 25 runs.}}
 \label{fig:repellers}
\end{figure}

\subsection{Control of a particle simulation}

We also applied these methods to optimize the behavior of a simulated physical system in which the trajectories of falling particles are controlled via a set of repelling forces. 
This is a difficult, nonlinear control task whose resulting objective function exhibits fairly isolated regions of high value surrounded by severe plateaus.
Briefly, the four-dimensional state-space in this problem consists of a particle's 2D position and velocity $(p,\dot p)$ with two-dimensional actions consisting of forces which act on the particle. Particles are also affected by gravity and a frictional force resisting movement. The goal is to direct the path of the particle through regions of the state space with high reward $r(p)$ so as to maximize the \emph{total reward} accumulated over many time-steps. In our experiments we use a finite, but large, time-horizon $H$. In order to control this system we employ a set of ``repellers'' each of which is located at some position $c_i=(a_i,b_i)$ and has strength $w_i$ (see the left-most plot of Figure~\ref{fig:repellers}). The force on a particle at position $p$ is a weighted sum of the individual forces from all repellers, each of which is inversely proportional to the distance $p-c_i$. For further details we refer the reader to \cite{hoffman-repellers}. 

This problem can be formulated in the setting of Bayesian optimization by defining the vector of repeller parameters $\x = (w_1,a_1,b_1,\dots)$. In the experiments shown in Figure~\ref{fig:repellers} we will utilize three repellers, resulting in a 9D optimization task. We can then define our objective as the total $H$-step expected reward $f(\x) = \mathbb E\big[\textstyle{\sum_{n=0}^H r(p_n)}\vert\x\big]$. Finally, since the model defines a probability distribution $P_\x(p_{0:H})$ over particle trajectories we can obtain a noisy estimate of this objective function by sampling a single trajectory and evaluating the sum over its immediate rewards.

Results for this optimization task are shown in Figure~\ref{fig:repellers}. As with the previous synthetic examples GP-Hedge outperforms each of its constituent methods. We further note the particularly poor performance of PI on this example, which in part we hypothesize is a result of plateaus in the resulting objective function. In particular PI has trouble exploring after it has ``locked on'' to a particular mode, a fact that seems exacerbated when there are large regions with very little change in objective.


\section{Convergence behaviour}

Properly assessing the convergence behaviour of hedging algorithms of this type is very problematic. The main difficulty lies with the fact that decisions made at iteration $t$ affect the state of the problem and the resulting rewards at all future iterations. As a result we cannot relate the regret of our algorithm directly to the regret of the \emph{best} underlying acquisition strategy: had we actually used the best underlying strategy we would have selected completely different points~\cite[section 7.11]{Cesa-Bianchi:2006}. 

Regret bounds for the underlying GP-UCB algorithm have been shown~\cite{Srinivas:2010}. Starting with Auer et al.\ we also have regret bounds for the hedging strategies used to select between acquisition functions~\cite{Auer:1998} (improved bounds can also be found in~\cite{Cesa-Bianchi:2006}). However, because of the points stated in the previous paragraph, and expounded in more detail in the appendix, we cannot simply combine both regret bounds.



With these caveats in mind we will consider a slightly different algorithmic framework. In particular we will consider rewards at iteration $t$ given by the mean $\mu_{t-1}(\x_t)$, where
this assumption is made merely to simplify the following proof. We will also assume that GP-UCB is included as one of the possible acquisition functions due to its associated convergence results (see Section~\ref{sec:acquisition}). In this scenario we can obtain the following bound on our cumulative regret.
\begin{theorem}
\label{theorem}
Assume GP-Hedge is used with a collection of acquisition strategies, one of which is GP-UCB with parameters $\beta_t$. If we also have a bound $\gamma_T$ on the information gained at points selected by the algorithm after $T$ iterations, then with probability at least $1-\delta$ the cumulative regret is bounded by
\begin{align*}
	R_T 
	&\leq \sqrt{TC_1\beta_T\gamma_T} + 
	\Big[\sum_{t=1}^T\beta_t\sigma_{t-1}(\x_t^\mathrm{UCB})\Big] + 
	\mathcal O(\sqrt{T}),
\end{align*}
where $\x_t^\mathrm{UCB}$ is the $t$th point proposed by GP-UCB.
\end{theorem}

We give a full proof of this theorem in the appendix. We will note that this theorem on its own does not guarantee the convergence of the algorithm, i.e.\ that $\lim_{T\to\infty}R_T/T=0$. We can see, however, that our regret is bounded by two sub-linear terms and an additional term which depends on the information gained at points proposed, but not necessarily selected. In some sense this additional term depends on the proximity of points proposed by GP-UCB to points previously selected, the expected distance of which should decrease as the number of iterations increases. 

\section{Conclusions and future work}

Hedging strategies are a powerful tool in the design of acquisition functions for Bayesian optimization. In this paper we have shown that strategies that adaptively modify a portfolio of acquisition functions often perform substantially better --- and almost never worse --- than the best-performing individual acquisition function. Our experiments have also shown that full-information strategies are able to outperform partial-information strategies in many situations. However, partial-information strategies can be beneficial in instances of high $N$ or in situations where the acquisition functions provide very conflicting advice. Evaluating these tradeoffs is an interesting area of future research.

Finally, while the EI and PI acquisition functions can perform well in practice, there currently exist no regret bounds for these approaches. In this work we give a regret bound for our hedging strategy by relating its performance to existing bounds for GP-UCB. Although our bound does not guarantee convergence it does provide some intuition as to the success of hedging methods in practice. Another interesting line of future research involves finding similar bounds for the gap measure.

\section*{Acknowledgements}

We would like to anonymously thank a number of researchers who provided very helpful comments and criticism on the theoretical and practical aspects of this work. 

\nocite{Grunewalder:2010}

{\small
\bibliographystyle{abbrv}
\bibliography{hedging}
}

\newpage
\appendix
\section{Proof of Theorem~\ref{theorem}}

We will consider a portfolio-based strategy using rewards $r_t=\mu_{t-1}(\x_t)$ and selecting between acquisition functions using the Hedge algorithm. In order to discuss this we will need to write the gain over $T$ steps, in hindsight, that would have been obtained had we used strategy $i$,
\begin{equation*}
	g_T^i = \sum_{t=1}^T r_t^i = \sum_{t=1}^T \mu_{t-1}(\x_t^i).
\end{equation*}
We must emphasize however that this gain is conditioned on the actual decisions made by Hedge, namely that points $\{\x_1,\dots,\x_{t-1}\}$ were selected by Hedge. If we define the maximum strategy $g_T^\mathrm{max}=\max_ig_T^i$ we can then bound the regret of Hedge with respect to this gain.
\begin{lemma}
\label{lemma:exp3}
With probability at least $1-\delta_1$ and for a suitable choice of Hedge parameters, $\eta=\sqrt{8\ln k/T}$, the regret is bounded by
\begin{equation*}
	g_T^\mathrm{max} - g_T^\mathrm{Hedge} \leq \mathcal O(\sqrt{T}).
\end{equation*}
\end{lemma}
This result is given without proof as it follows directly from \cite[Section 4.2]{Cesa-Bianchi:2006} for rewards in the range $[0,1]$. At the cost of slightly worsening the bound in terms of its multiplicative/additive constants, the following generalizations can also be noted:
\setlength\plparsep{5pt}
\begin{compactitem}
	\item For rewards instead in the arbitrary range\footnote{
To obtain rewards bounded within some range $[a,b]$ we can assume that the additive noise $\epsilon_t$ is truncated above some large absolute value, which guarantees bounded means.} $[a,b]$ the same bound can be shown by referring to \cite[Section 2.6]{Cesa-Bianchi:2006}.

	\item The choice of $\eta$ in the above Lemma requires knowledge of the time horizon $T$. By referring to \cite[Section 2.3]{Cesa-Bianchi:2006} we can remove this restriction using a time-varying term $\eta_t=\sqrt{8\ln k/t}$.
	
	\item By referring to \cite[Section 6.8]{Cesa-Bianchi:2006} we can also extend this bound to the partial-information strategy Exp3.
\end{compactitem}
Finally, we should also note that this regret bound trivially holds for any strategy $i$, since $g_T^\mathrm{max}$ is the maximum. It is also important to note that this lemma holds for any choice of $r_t^i$, \emph{with rewards depending on the actual actions taken by Hedge}. The particular choice of rewards we use for this proof have been selected in order to achieve the following derivations.

For the next two lemmas we will refer the reader to \cite[Lemma 5.1 and 5.3]{Srinivas:2010} for proof. 
We point out, however, that these two lemmas only depend on the underlying Gaussian process and as a result can be used separately from the GP-UCB framework.
\begin{lemma}
	\label{lemma:concentration}
	Assume $\delta_2\in(0,1)$, a finite sample space $|A|<\infty$, and $\beta_t=2\log(|A|\pi_t/\delta_2)$ where $\sum_{t}\pi_t^{-1}=1$ and $\pi_t>0$. Then with probability at least $1-\delta_2$ the absolute deviation of the mean is bounded by
	\begin{equation*}
		|f(\x) - \mu_{t-1}(\x)| \leq \sqrt{\beta_t}\sigma_{t-1}(\x)
		\quad\forall x\in A, \forall t\geq 1.
	\end{equation*}
\end{lemma}
In order to simplify this discussion we have assumed that the sample space $A$ is finite, however this can also be extended to compact spaces \cite[Lemma 5.7]{Srinivas:2010}.
\begin{lemma}
	\label{lemma:info}
	The information gain for points selected by the algorithm can be written as
	\begin{equation*}
		I(y_{1:T}; f_{1:T}) 
		= \frac12 
		\sum_{t=1}^T\log(1+\sigman^{-2}\sigma^2_{t-1}(\x_t)).
	\end{equation*}
\end{lemma}

The following lemma follows the proof of \cite[Lemma 5.4]{Srinivas:2010}, however it can be applied outside the GP-UCB framework. Due to the slightly different conditions we recreate this proof here.
\begin{lemma}
	\label{lemma:sum}
	Given points $x_t$ selected by the algorithm the following bound holds for the sum of variances:
	\begin{equation*}
		\sum_{t=1}^T \beta_t \sigma_t^2(\x_t) \leq C_1\beta_T\gamma_T,
	\end{equation*}
	where $C_1=2/\log(1+\sigman^{-2})$.
\end{lemma}
\begin{proof}
	Because $\beta_t$ is nondecreasing we can write the following inequality
	\begin{align*}
		\beta_t\sigma_{t-1}^2(\x_t)
		&\leq
		\beta_T\sigman^2(\sigman^{-2}\sigma_{t-1}^2(\x_t)) \\
		&\leq
		\beta_T\sigman^2
		\frac{\sigman^{-2}}{\log(1+\sigman^{-2})}
		\log(1+\sigman^{-2}\sigma^2_{t-1}(\x_t)).
	\end{align*}
	The second inequality holds because the posterior variance is bounded by the prior variance, $\sigma_{t-1}^2(\x)\leq k(\x,\x)\leq 1$, which allows us to write
	\begin{align*}
		\sigman^{-2}\sigma^2_{t-1}(\x_t) 
		&\leq
		\sigman^{-2}
		\frac
		{\log(1+\sigman^{-2} \sigma_{t-1}^2(\x_t))}
		{\log(1+\sigman^{-2})\hphantom{\sigma_{t-1}^2(\x_t)}} .
	\end{align*}
	By summing over both sides of the original bound and applying the result of Lemma~\ref{lemma:info} we can see that
	\begin{align*}
		\sum_{t=1}^T \beta_t\sigma_{t-1}^2(\x_t)
		&\leq 
		\beta_T \frac12 C_1 
		\sum_{t=1}^T \log(1+\sigman^{-2}\sigma^2_{t-1}(\x_t)) \\
		&= \beta_T C_1 I(y_{1:T}; f_{1:T}).
	\end{align*}
	The result follows by bounding the information gain by $I(y_{1:T}; f_{1:T})\leq \gamma_T$, which can be done for many common kernels, including the squared exponential \cite[Theorem 5]{Srinivas:2010}.
\end{proof}

Finally, the next lemma follows directly from \cite[Lemma 5.2]{Srinivas:2010}. We will note that this lemma depends only on the definition of the GP-UCB acquisition function, and as a result does not require that points at any previous iteration were actually selected via GP-UCB.
\begin{lemma}
	\label{lemma:ucb}
	If the bound from Lemma~\ref{lemma:concentration} holds, then for a point $\x_t^\mathrm{UCB}$ proposed by GP-UCB with parameters $\beta_t$ the following bound holds:
	\begin{equation*}
		f(\x^*) - \mu_{t-1}(\x_t^\mathrm{UCB}) \leq 
		\sqrt{\beta_t}\sigma_{t-1}(\x_t^\mathrm{UCB}).
	\end{equation*}
\end{lemma}

We can now combine these results to construct the proof of Theorem~\ref{theorem}.
\begin{proof}[Proof of Theorem~\ref{theorem}]
With probability at least $1-\delta_1$ the result of Lemma~\ref{lemma:exp3} holds. If we assume that GP-UCB is included as one of the acquisition functions we can write
\begin{equation*}
	-g_T^\mathrm{Hedge} \leq \mathcal O(\sqrt{T}) - g_T^\mathrm{UCB}
\end{equation*}
and by adding $\sum_{t=1}^T f(\x^*)$ to both sides this inequality can be rewritten as
\begin{equation*}
	\sum_{t=1}^T f(\x^*) - \mu_{t-1}(\x_t) \leq \mathcal O(\sqrt{T}) + 
	\sum_{t=1}^T f(\x^*) - \mu_{t-1}(\x_t^\mathrm{UCB}).
\end{equation*}
With probability at least $1-\delta_2$ the bound from Lemma~\ref{lemma:concentration} can be applied to the left-hand-side and the result of Lemma~\ref{lemma:ucb} can be applied to the right. This allows us to rewrite this inequality as
\begin{align*}
	&\sum_{t=1}^T f(\x^*) - f(\x_t) - \sqrt{\beta_t}\sigma_{t-1}(\x_t) \\
	&\hspace{3em}
	\leq
	\mathcal O(\sqrt{T}) + 
	\sum_{t=1}^T \sqrt{\beta_t}\sigma_{t-1}(\x_t^\mathrm{UCB})
\end{align*}
which means that the regret is bounded by
\begin{align*}
	R_T
	&= \sum_{t=1}^T f(\x^*) - f(\x_t) \\
	&\leq
	\mathcal O(\sqrt{T}) + 
	\sum_{t=1}^T \sqrt{\beta_t}\sigma_{t-1}(\x_t^\mathrm{UCB}) + 
	\sum_{t=1}^T \sqrt{\beta_t}\sigma_{t-1}(\x_t) \\
	&\leq
	\mathcal O(\sqrt{T}) + 
	\sum_{t=1}^T \sqrt{\beta_t}\sigma_{t-1}(\x_t^\mathrm{UCB}) + 
	\sqrt{C_1T\beta_T\gamma_T}.
\end{align*}
This final inequality follows directly from Lemma~\ref{lemma:sum} by application of the Cauchy-Schwarz inequality. We should note that we cannot use Lemma~\ref{lemma:sum} to further simplify the terms involving the sum over $\sigma_{t-1}(\x_t^\mathrm{UCB})$. This is because the lemma only holds for points that are sampled by the algorithm, which may not include those proposed by GP-UCB.

Finally, this result depends upon Lemmas~\ref{lemma:exp3} and~\ref{lemma:ucb} holding. By a simple union bound argument we can see that these both hold with probability at least $1-\delta_1-\delta_2$, and by setting $\delta_1=\delta_2=\delta/2$ we recover our result.
\end{proof}

\section{Synthetic test functions}\label{sec:synthetics}

As there is no generally-agreed-upon set of test functions for Bayesian optimization in higher dimensions, we seek to sample synthetic functions from a known GP prior, similar to the strategy of Lizotte~\cite{Lizotte:2008}.  A GP prior is infinite-dimensional, so on a practical level for performing experiments we simulate this by sampling points and using the posterior mean as the synthetic objective test function.  

For each trial, we use an ARD kernel with $\thetav$ drawn uniformly from $[0,2]^d$.  We then sample $100d$ $d$-dimensional points, compute $\K$ and then draw $\y \sim \mathcal{N}(0, \K)$.   The posterior mean of the resulting predictive posterior distribution $\mu(\x)$ (Section~\ref{sec:gp}) is used as the test function.  However it is possible that for particular values of $\thetav$ and $\K$, large parts of the space will be so far from the samples that they will form plateaus along the prior mean.  To reduce this, we evaluate the test function at 500 random locations.  If more than 25 of these are 0, we recompute $\K$ using $200d$ points.  This process is repeated, adding $100d$ points each time until the test function passes the plateau test (this is rarely necessary in practice).

Using the response surface $\mu(\x)$ as the objective function, we can then approximate the maximum using conventional global optimization techniques to get $f(\x^\star)$, which permits us to use the gap metric for performance.  

Note that these sample points are only used to construct the objective, and are not known to the optimization methods.

\end{document}